\documentclass[sigconf]{acmart}
\AtBeginDocument{%
  \providecommand\BibTeX{{%
    \normalfont B\kern-0.5em{\scshape i\kern-0.25em b}\kern-0.8em\TeX}}}

\usepackage{multirow}
\usepackage{multicol}
\usepackage{graphicx}
\usepackage{algorithm}
\usepackage{algpseudocode}  
\usepackage{amsmath} 
\usepackage{subcaption}
\usepackage{makecell}

\usepackage{bbm}

\renewcommand{\algorithmiccomment}[1]{\textbf{Code:}}

\usepackage{enumitem}
\setlist[itemize,1]{leftmargin=\dimexpr 26pt-2mm}

\copyrightyear{2022} 
\acmYear{2022} 
\setcopyright{acmcopyright}
\acmConference[KDD '22]{Proceedings of the 28th ACM SIGKDD Conference on Knowledge Discovery and Data Mining}{August 14--18, 2022}{Washington, DC, USA}
\acmBooktitle{Proceedings of the 28th ACM SIGKDD Conference on Knowledge Discovery and Data Mining (KDD '22), August 14--18, 2022, Washington, DC, USA}
\acmPrice{15.00}
\acmDOI{10.1145/3534678.3539021}
\acmISBN{978-1-4503-9385-0/22/08}

\begin{document}

% \title{POLARIS: A Geographic Pre-trained Model and its Applications in Baidu Maps}
\title{ERNIE-GeoL: A Geography-and-Language Pre-trained Model and its Applications in Baidu Maps}

\author{Jizhou Huang,
	Haifeng Wang, 
	Yibo Sun,
	Yunsheng Shi,
	Zhengjie Huang,
	An Zhuo,
  Shikun Feng}
\thanks{$^*$Corresponding author: Jizhou Huang.}
\affiliation{%
\institution{Baidu Inc., China}
\country{}
}

\email{{huangjizhou01, wanghaifeng, sunyibo, shiyunsheng01, huangzhengjie, zhuoan, fengshikun01}@baidu.com}

\renewcommand{\shortauthors}{Jizhou Huang et al.}


\begin{abstract}
Pre-trained models (PTMs) have become a fundamental backbone for downstream tasks in natural language processing and computer vision.
Despite initial gains that were obtained by applying generic PTMs to geo-related tasks at Baidu Maps, a clear performance plateau over time was observed.
One of the main reasons for this plateau is the lack of readily available geographic knowledge in generic PTMs.
To address this problem, in this paper, we present ERNIE-GeoL, which is a geography-and-language pre-trained model designed and developed for improving the geo-related tasks at Baidu Maps.
ERNIE-GeoL is elaborately designed to learn a universal representation of geography-language by pre-training on large-scale data generated from a heterogeneous graph that contains abundant geographic knowledge.
Extensive quantitative and qualitative experiments conducted on large-scale real-world datasets demonstrate the superiority and effectiveness of ERNIE-GeoL.
ERNIE-GeoL has already been deployed in production at Baidu Maps since April 2021, which significantly benefits the performance of various downstream tasks.
This demonstrates that ERNIE-GeoL can serve as a fundamental backbone for a wide range of geo-related tasks.
\end{abstract}

\begin{CCSXML}
<ccs2012>
<concept>
<concept_id>10002951.10003227.10003351</concept_id>
<concept_desc>Information systems~Data mining</concept_desc>
<concept_significance>500</concept_significance>
</concept>
</ccs2012>
\end{CCSXML}

\ccsdesc[500]{Information systems~Data mining}
\keywords{pre-training, heterogeneous graph, graph neural network}
% \keywords{pre-training, heterogeneous graph, graph neural network, Baidu Maps}


\maketitle

\section{Introduction}
\label{sec:intro}

\begin{figure}[t!]
\setlength{\abovecaptionskip}{0.15cm}
\setlength{\belowcaptionskip}{0.01cm}
\centering
\includegraphics[width=\columnwidth,trim={0.1cm 0.1cm 0.1cm 0.1cm},clip]{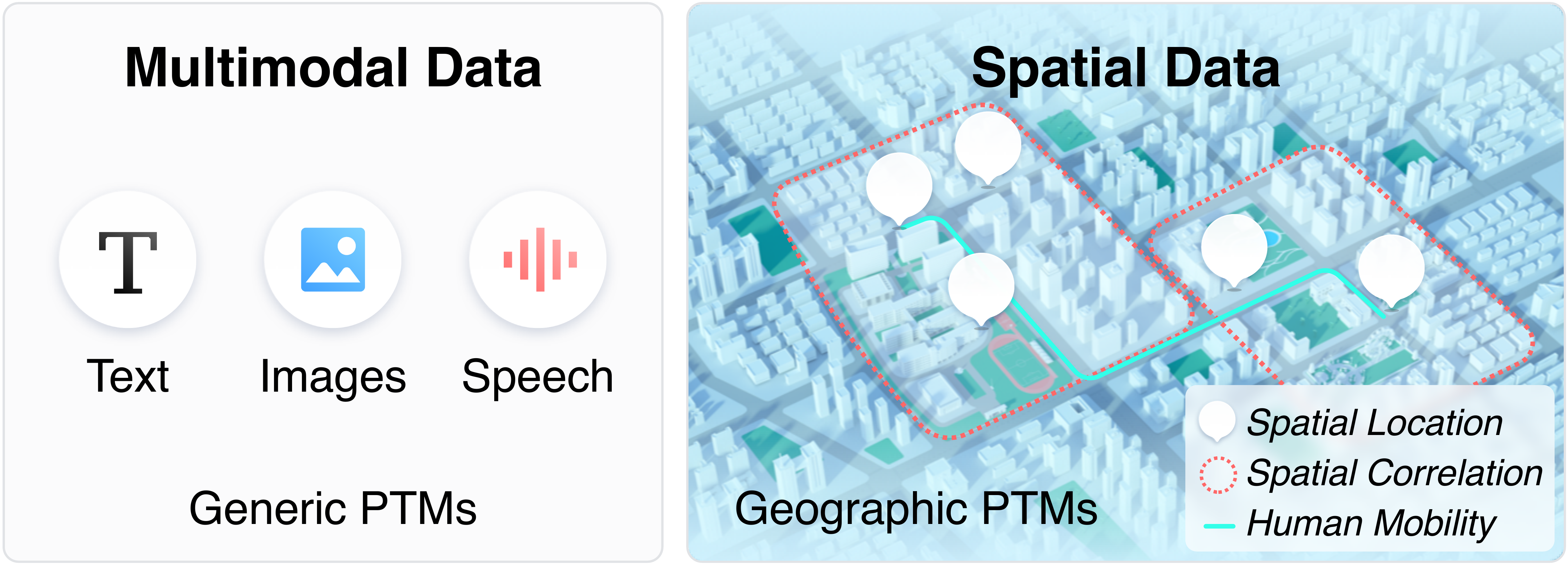}
\caption{Illustration of the multimodal data and spatial data used for training generic and geographic PTMs, respectively.}
\label{fig:show_case}
\vspace{-6mm}
\end{figure}

Pre-trained models (PTMs) are designed to learn a universal representation from large-scale raw text \cite{peters-etal-2018-deep,Devlin2019BERTPO,sun2020ernie}, unlabeled images \cite{lu2019vilbert}, or videos \cite{sun2019videobert}, which have become a fundamental backbone for downstream tasks in natural language processing (NLP) and computer vision (CV)~\cite{bommasani2021opportunities}.
The most common paradigm for adapting PTMs to downstream tasks is sequential transfer learning \cite{pan2009survey} via supervised fine-tuning on labeled data.
In this paradigm, downstream tasks can benefit from the knowledge learned by PTMs, which brings significant improvements \cite{Devlin2019BERTPO}.

The web mapping services provided by Baidu Maps, such as point of interest (POI) retrieval~\cite{hgamn-kdd21,p3ac-kdd20,metasug-kdd21}, POI recommendation~\cite{metapoirec-kdd21}, POI information page~\cite{gedit-cikm21}, and intelligent voice assistant~\cite{huang2022DuIVA}, have shown improved performance by applying PTMs.
However, a clear performance plateau over time was observed in our practice, i.e., the performance gain remains marginal w.r.t. the optimization of generic PTMs.
One of the main reasons for this plateau is the lack of geographic knowledge, which plays a vital role in improving tasks that necessitate computational support for geographic information (hereafter referred to as geo-related tasks).
In this work, we focus on two types of geographic knowledge.
(1) \textbf{Toponym knowledge.} A toponym refers to the name of a geo-located entity, such as a POI, a street, and a district. Toponym resolution \cite{10.1145/1328964.1328989}, which aims at identifying and extracting toponyms from text, is a fundamental  necessity for a wide range of geo-related tasks. However, the semantic meaning of most toponyms can hardly be captured by generic PTMs, because toponym knowledge is largely absent from or rarely seen in their training data. 
(2) \textbf{Spatial knowledge.} Spatial knowledge mainly includes the geographic coordinates of a geo-located entity and the spatial relationships between different geo-located entities, which is indispensable for geo-related tasks such as geocoding~\cite{goldberg2007text} and georeferencing \cite{hill2009georeferencing}.
However, the generic PTMs are incapable of handling geo-related tasks effectively, due to the absence of spatial knowledge and the lack of pre-training tasks for incorporating spatial knowledge.

To effectively learn a geography-and-language pre-trained model from large-scale spatial data, we need to address two key challenges. (1) \textbf{Heterogeneous data integration.}
Figure~\ref{fig:show_case} illustrates the main difference between the training data used by generic and geographic PTMs.
Generic PTMs are typically learned from multimodal data, including text, images, and speech.
By contrast, the spatial data mainly include spatial location (a single POI), spatial correlation (a triplet of POIs), and human mobility (a sequence of POIs).
However, the way to effectively integrate the three sources of spatial data with text data for training geographic PTMs has been little explored and remains a challenge.
(2) \textbf{Geography-language pre-training.} Different from existing language model pre-training \cite{Devlin2019BERTPO} and vision-language (image-text \cite{lu2019vilbert} and video-text \cite{sun2019videobert}) pre-training that is designed to learn the semantic correlations between vision and language, geography-language pre-training necessitates learning the associations between geography and language, e.g., learning to associate ``Beijing railway station'' in text with its real-world geo-location information in the form of coordinates.
To learn a cooperative knowledge of geography and language from unlabeled data,
it is important to design effective backbone networks, pre-training tasks, and learning objectives.

In this paper, we present our efforts toward designing and implementing ERNIE-GeoL, which is a geography-and-language pre-trained model designed for improving a wide range of geo-related downstream tasks at Baidu Maps.
Specifically, we \textbf{first} construct a heterogeneous graph that contains POI nodes and query nodes, using the POI database and search logs of Baidu Maps. 
To integrate spatial information with text, we construct edges between two nodes based on spatial correlation and human mobility data, as shown in Figure \ref{fig:show_case}, which enables knowledge transfer between different modalities.
To generate each input sequence for training ERNIE-GeoL, we use the random walk algorithm to sample a sequence of nodes as an input document.
In this way, we can automatically build large-scale training data, which facilitates comprehensive knowledge transfer and bridges the modality gap.
\textbf{Second}, we use Transformer as the backbone network to learn the representations of each node.
To incorporate an input document's graph information, we employ a Transformer-based aggregation layer to encode the relations between multiple nodes in the document.
To effectively learn comprehensive knowledge, we adopt masked language modeling and geocoding as the pre-training tasks, which are elaborated to simultaneously learn toponym and spatial knowledge, as well as to balance both knowledge explorations.
As such, we can learn a universal representation of geography-language by pre-training on domain-specific and cross-modality data.

We evaluate ERNIE-GeoL on five geo-related tasks.
Extensive experiments show that ERNIE-GeoL significantly outperforms the generic PTMs when applied to all five tasks.
ERNIE-GeoL has already been deployed in production at Baidu Maps since April 2021, which significantly benefits the performance of a wide range of downstream tasks. This demonstrates that ERNIE-GeoL can serve as a fundamental backbone for geo-related tasks.

Our contributions can be summarized as follows:
\begin{itemize}[topsep=3pt]
    \item \textbf{Potential impact}: We suggest a practical and robust solution for training a geography-and-language pre-trained model, named ERNIE-GeoL, which can serve as a fundamental backbone for geo-related tasks. We document our efforts and findings on designing and developing ERNIE-GeoL, and we hope that it could be of potential interest to practitioners working with pre-trained models and geo-related problems. 
    \item \textbf{Novelty}: The design and development of ERNIE-GeoL are driven by the novel idea that learns a universal representation of geography-language by pre-training on large-scale graph data with both toponym and spatial knowledge. To the best of our knowledge, this is the first attempt to design and build a geography-and-language pre-trained model.
    \item \textbf{Technical quality}: Extensive quantitative and qualitative experiments, conducted on large-scale, real-world datasets, demonstrate the superiority and effectiveness of ERNIE-GeoL. The successful deployment of ERNIE-GeoL at Baidu Maps further shows that it is a practical and fundamental backbone for a wide range of geo-related tasks.
\end{itemize}

\section{ERNIE-G\texorpdfstring{\MakeLowercase{eo}}{eo}L}
In this section, we introduce the design and implementation details of ERNIE-GeoL, which mainly contains three parts: training data construction, model architecture, and pre-training tasks.

\label{sec:methods}
\begin{figure*}[ht]
\setlength{\abovecaptionskip}{0.25cm}
\centering
\includegraphics[width=\textwidth,trim={0.05cm 0.05cm 0.05cm 0.1cm},clip]{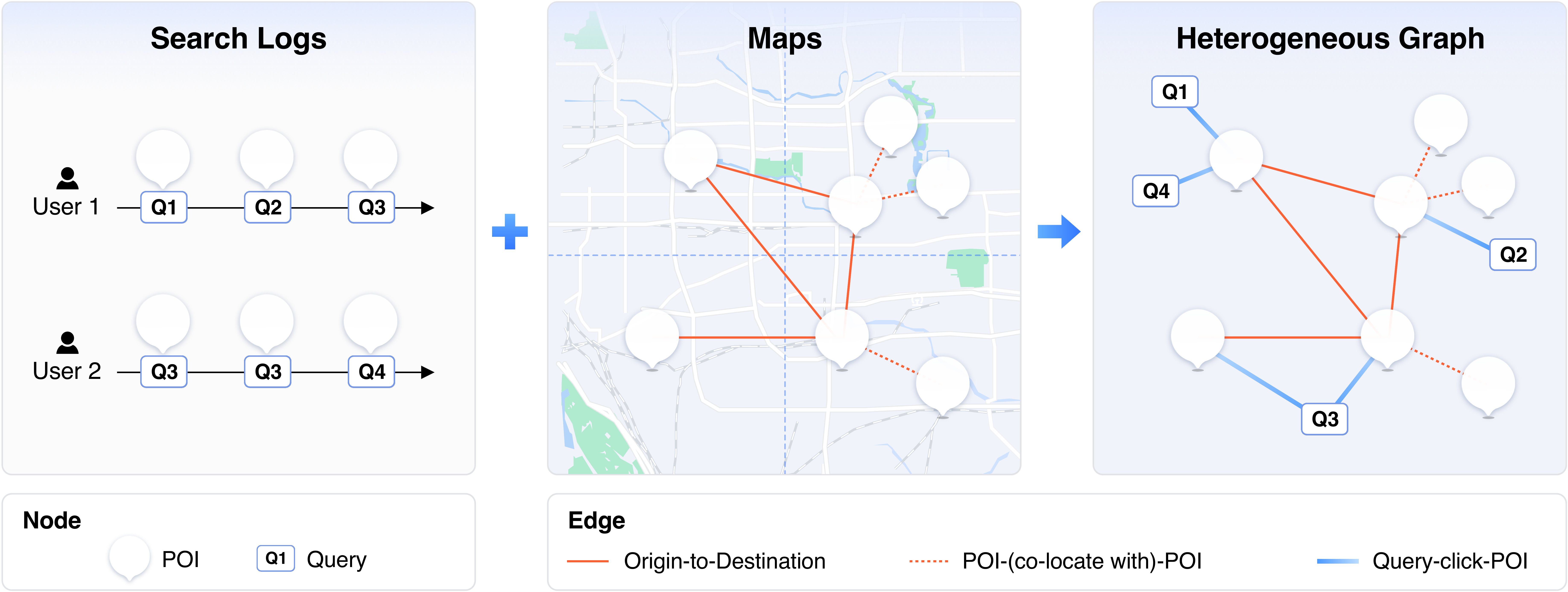}
\caption{The process of constructing the heterogeneous graph.}
\label{fig:construct_graph}
\vspace{-3mm}
\end{figure*}

\subsection{Training Data Construction}
\label{sec:data_construction}
Our previous work \cite{hgamn-kdd21} has demonstrated that the heterogeneous graph is able to significantly benefit POI retrieval task.
Motivated by this, we construct large-scale training data based on a heterogeneous graph that contains both toponym and spatial knowledge for pre-training ERNIE-GeoL.
Specifically, we first construct a unified heterogeneous graph that contains POI nodes and query nodes using the POI database and search logs of Baidu Maps.
Then, we construct edges between two nodes based on spatial relationships between POIs to integrate spatial information with text.
The toponym data mainly include POI names and addresses, which are derived from the POI database and are stored in unstructured text format. 
The spatial data consist of POI geographic coordinates, POIs that co-locate within individual geographic regions, and POIs that co-occur in same sessions from search logs, which are stored using numerical digits or in triplet format (i.e., non-text format).

To bridge the gap between the text and non-text representations, we build a heterogeneous graph $\mathcal{G} = (\mathcal{V}, \mathcal{E}, O_\mathcal{V}, R_\mathcal{E})$, where $\mathcal{V}$ denotes the set of nodes, $\mathcal{E}$ the set of edges, $O_\mathcal{V}$ the set of node types, and $R_\mathcal{E}$ the set of edge types. Each node $v \in \mathcal{V}$ and each edge $e \in \mathcal{E}$ are associated with their corresponding mapping functions $\psi(v):\mathcal{V} \rightarrow O_\mathcal{V} $ and $ \phi(e): \mathcal{E} \rightarrow R_\mathcal{E}$.
As shown by Figure~\ref{fig:construct_graph}, the node types $O_\mathcal{V}$ include \textit{POI} and \textit{query}.
The edge types $R_\mathcal{E}$ include \textit{Query-click-POI}, \textit{Origin-to-Destination}, and \textit{POI-(co-locate with)-POI}.
Next, we detail each element of the heterogeneous graph.

\subsubsection{POI Node and Query Node}
\label{sec:node_text_repr}
Assigning a unique ID to each POI node is a straightforward way to represent it.
However, this approach is unable to represent newly emerging POIs.
To address this problem, we uniformly use text rather than fixed IDs to represent all nodes.
% each query node is the text used to search a user's desired POI.
% Each POI node represents a POI in the POI database.
Specifically, we organize each POI node in the form of the concatenation of the following three types of textual information: (1) the full POI name, (2) the POI address, and (3) the POI type.
We separate each type of textual information with a special token ``[SEP]''.
We also equip each POI node with its real-world location information, i.e., the geographic coordinates of it.
See Appendix \ref{appendix:qp_repr} for a detailed description of the node representation method.

\subsubsection{Query-click-POI Edge}
After typing in a query, a user would click on the desired POI from a list of ranked POIs that the POI search engine suggested.
This process produces large-scale query-POI pairs, where the different expressions of each POI can bridge the semantic gap between queries and POIs.
For example, users usually make spelling errors or use abbreviations, which would lead to poor results when directly matching query and POI text information.
Motivated by this observation, we model the relations between queries and POIs using the Query-click-POI edge.
Specifically, we connect Query-click-POI edges between each POI node and its historical query nodes.
To speed up training with large-scale data, we select the top 4 searched queries for each POI by following \cite{hgamn-kdd21}.

\subsubsection{Origin-to-Destination Edge}
A user's mobility behavior \cite{10.1145/3394486.3412856,xiao2021c,huang2020quantifying} produces a visited POI sequence in search logs.
From which, the origin POI and destination POI can be extracted to construct the Origin-to-Destination edge between two POIs.
Specifically, we perform a 2-gram sliding window on the POI sequences and link the adjacent POIs with such edge type.

\subsubsection{POI-(co-locate with)-POI Edge}
POIs that co-locate within individual geographic regions may exhibit a high degree of spatial similarity and share common features.
Motivated by this observation, we build an additional spatial relations between different POIs by introducing the POI-(co-locate with)-POI edge.
As shown in the middle of Figure~\ref{fig:construct_graph}, we quantize the Earth's surface as a grid using a Discrete Global Grid\footnote{\url{https://en.wikipedia.org/wiki/Discrete_global_grid}} (DGG) system and construct the co-location edge between POIs that lie in the same cell of the grid.
This can be regarded as an analogy to the surrounding words, where each word in a sentence is surrounded by a textual context.
In this way, each POI node in the heterogeneous graph has a spatial context representing the real-world spatial distribution.

Specifically, we use the S2 geometry\footnote{\url{https://s2geometry.io}} as the DGG system.
See Appendix \ref{appendix:geocode_system} for a detailed comparison of recently proposed DGG systems and the reason that we choose S2 geometry.
The S2 geometry library supports 31 levels of hierarchy, where its grid's cells have different area coverage.
For simplicity, we use the term ``S2 cell'' to refer to the cells generated by the S2 geometry library.
We construct the co-location edges between the POIs within the same S2 cell of level 15, which covers an area of \textasciitilde 200 m$^2$.

\begin{center}
\setlength{\textfloatsep}{1pt} % set a smaller \textfloatsep 
\begin{algorithm}[!t]
\caption{Random Walk Sampling on Heterogeneous Graph} 
\label{code:recentEnd} 
\begin{algorithmic}[1] 
\Require 
Heterogeneous Graph $\mathcal{G} = (\mathcal{V}, \mathcal{E}, O_\mathcal{V}, R_\mathcal{E})$; $n$ walk length
\Ensure 
Sequence of text nodes $D=\{v_1, v_2,\cdots, v_n\}$
\item[]
\For{each $v \in \mathcal{G}$} 
    \State $D = [v, ]$
    \For{$i=0$ to $n$}
        \State draw u according to the Equation \ref{eq:rw}.
        \State $D.push(u)$
        \State $v=u$
    \EndFor 
\EndFor
\end{algorithmic} 
\end{algorithm}
\end{center}

\subsubsection{Random Walk Sampling}
\label{sec:random_walk}
Upon $\mathcal{G}$, we use the random walk algorithm to sample a sequence of nodes as an input document $D = \{v_1, v_2,\cdots, v_n\}$, where $n$ is the length of $D$.
Each node $v_i$ has a text representation consisting of a sequence of words $W_i = \{s^i_1, \cdots, s^i_j, \cdots, s^i_L\}$. During random walk sampling, at time step $i$, we sample the node by considering the influence of different edges. Specifically, we use the weighted probability distributed over the neighbors of $v_i$ as the transition probability $p(u|v)$:
\begin{equation}
\resizebox{1.0\columnwidth}{!}{$%
p(u|v)=\left\{
\begin{aligned}
\lambda_1 * \frac{1}{|N(e_{v,u})|} & , & \forall \phi(e_{v,u}) \in \text{Query-click-POI} \; edge \\
\lambda_2 * \frac{1}{|N(e_{v,u})|} & , & \forall \phi(e_{v,u}) \in \text{Origin-to-Destination} \; edge \\
\lambda_3 * \frac{1}{|N(e_{v,u})|} & , & \forall \phi(e_{v,u}) \in \text{POI-(co-locate with)-POI} \; edge \\
\end{aligned}
\right.
$%
}%
\label{eq:rw}
\end{equation}
where $|N(e_{v,u})|$ denotes the number of $v$'s neighborhood with corresponding edge, and $\lambda_i$ is the weight corresponding to different edges. Algorithm \ref{code:recentEnd} shows the details.
See Appendix \ref{appendix:data} for an illustration of the generated training examples.

\begin{figure*}[th!]
\setlength{\abovecaptionskip}{0.25cm}
	\centering
	\includegraphics[width=\textwidth,clip]{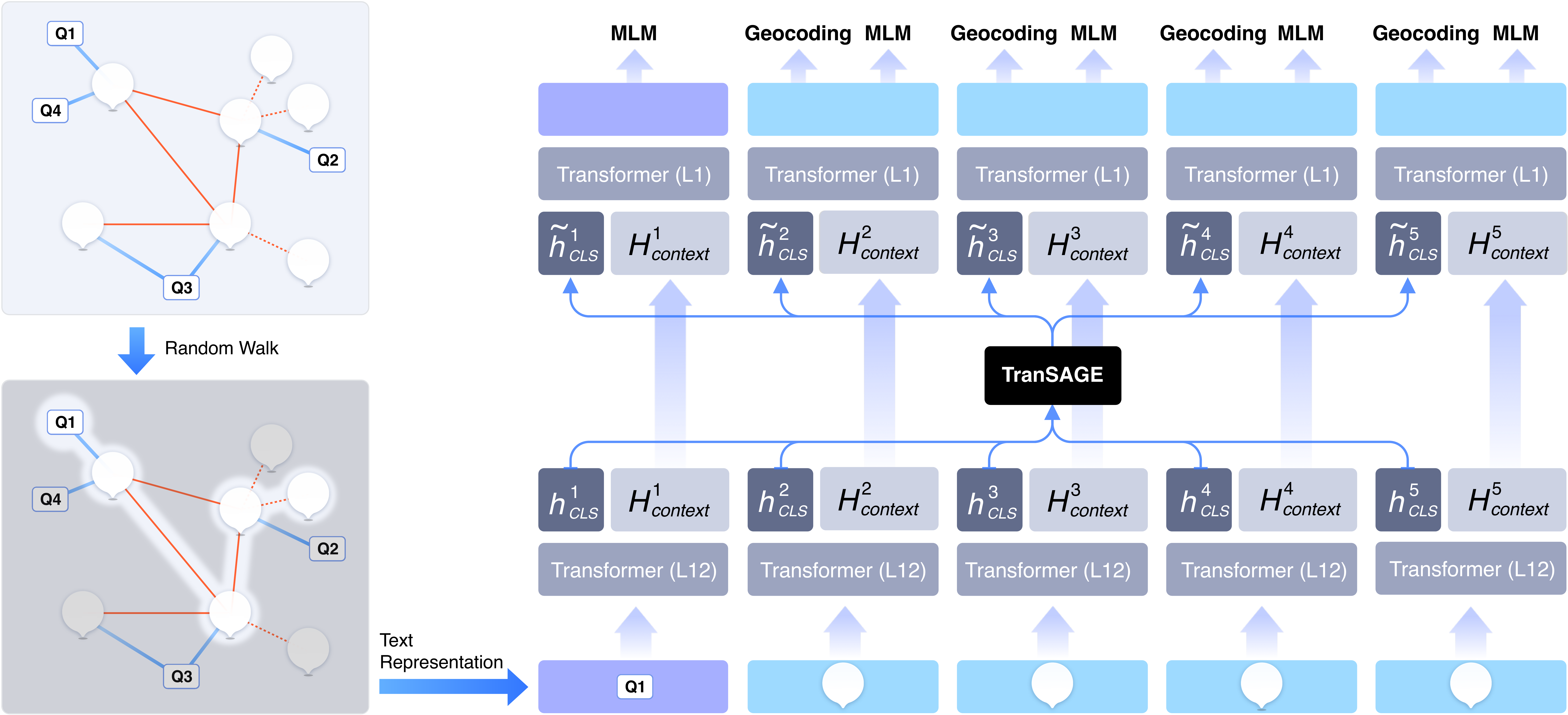}
	\caption{The architecture of ERNIE-GeoL.}
	\label{fig:model_structure}
\vspace{-3mm}
\end{figure*}

\subsection{Model Architecture}
As shown in Figure~\ref{fig:model_structure}, the two major components in ERNIE-GeoL's model architecture are a multi-layer bidirectional Transformer~\cite{vaswani2017attention} encoder and a \textbf{trans}former-based \textbf{ag}gr\textbf{e}gation (TranSAGE) layer.
The document in traditional NLP tasks consists of multiple sentences, where discourse structure of the document should be considered.
By contrast, the input document $D$ for pre-training ERNIE-GeoL consists of a sequence of nodes, and there is no discourse structure in $D$.
Therefore, instead of concatenating all the nodes and modeling them as one text sequence, we use the Transformer encoder to get each node's hidden vector separately, and employ the TranSAGE layer to capture the relations between each node and its neighbors.
As a graph-contextualized representation, the output of the TranSAGE is fused with each node's vector.
Then, each node's fused representation is used for the pre-training tasks.

Formally, for each node $v_i$ in document $D$, we first use the sentence-piece algorithm to tokenize its text representation $W_i$ into a sub-word sequence $S_i = \{s^i_1, \cdots, s^i_j, \cdots, s^i_L\}$, where $s^i_j$ denotes a sub-word token and $L$ is the length of the sub-word sequence.
Then, we insert a ``[CLS]'' token to the head of the tokenized sequence and use a multi-layer bidirectional Transformer encoder to get the node $v_i$'s vector representation $\{\mathbf{h}^i_{CLS}, \mathbf{H}^i_{context}\}$ as follows:
\begin{equation}
\begin{split}
\{\mathbf{h}^i_{CLS}, \mathbf{H}^i_{context}\} = \text{Transformer}(\{[CLS], s^i_1,\cdots,s^i_j,\cdots,s^i_L\}) ,
\end{split}
\label{equ:1}
\end{equation}
where $\mathbf{H}^i_{context}= \{\mathbf{h}^i_0, \cdots, \mathbf{h}^i_j, \cdots, \mathbf{h}^i_L\}$ and $\mathbf{h}^i_j \in \mathbb{R}^{d_h}$ is the hidden vector of $s^i_j$.
$\mathbf{h}^i_{CLS} \in \mathbb{R}^{d_h}$ is the final hidden vector of the special token ``[CLS]'', which is the aggregated representation of $S_i$.

After we obtain each node's vector representation, we need to model the relations between different nodes.
A straightforward way to model such relations is directly adopting the vanilla multi-head attention~\cite{vaswani2017attention} mechanism to project the input sequence into a key matrix and a value matrix with corresponding weights. The attention scores are computed by the dot product of these two matrices.
However, such attention mechanism is agnostic of the different node's type in the heterogeneous graph.
Taking node type into consideration, we propose a TranSAGE layer, which use different weights for the projection based on the node's type.
Specifically, we first pack the aggregated representations of all nodes together into a matrix $\mathbf{H} = \{\mathbf{h}^1_{CLS}, \cdots, \mathbf{h}^i_{CLS}, \cdots, \mathbf{h}^n_{CLS}\}$.
Then, we use the TranSAGE layer to compute the output matrix $\tilde{\mathbf{H}}$ as follows:
\begin{equation}
\begin{split}
    & \tilde{\mathbf{H}} = \mathbf{concat}(head_1, \cdots, head_l)W^O , \\
    & head_j = \text{softmax}(\frac{Q_jK^T_j}{\sqrt{d}})\mathbf{H} , \\
    & Q_j = \text{Q-Linear}_{\psi(v_i)}(\mathbf{H}) , \\
    & K_j = \text{K-Linear}_{\psi(v_i)}(\mathbf{H}) , \\
    \end{split}
    \end{equation}
where $\text{Q-Linear}_{\psi(v_i)} : \mathbb{R}^{d_h} \rightarrow \frac{\mathbb{R}^{d_h}}{l}$ is a linear projection indexed by each node's type, and $l$ is the number of the head.
We use this projection layer to convert $\mathbf{H}$ into a query matrix $Q_j$ for the $j$-th head, where nodes with different types are computed with unique parameters.
Similarly, we use another linear projection $\text{K-Linear}_{\psi(v_i)}$ to compute a key matrix $K_j$ for the $j$-th head.

Finally, we apply another attention-based module to each sub-word representation $\mathbf{h}^i_j$ with its corresponding graph contextualized representation $\tilde{\mathbf{h}}^i_{CLS} \in \tilde{\mathbf{H}}$.
Specifically, as shown in the right part of Figure~\ref{fig:model_structure}, we replace $\mathbf{h}^i_{CLS}$ in Equation~\ref{equ:1} with $\tilde{\mathbf{h}}^i_{CLS}$ and use a new Transformer layer to get each node's representation for computing the pre-training objectives as follows:
\begin{equation}
    \begin{split}
    \{\widehat{\mathbf{h}}^i_{CLS}, \widehat{\mathbf{H}}^i_{context}\} = \text{Transformer}(\{\tilde{\mathbf{h}}^i_{CLS}, \mathbf{H}^i_{context}\}) ,
    \end{split}
    \label{equ:2}
\end{equation}
where $\widehat{\mathbf{h}}^i_{CLS}$ is used for training the geocoding task and $\widehat{\mathbf{H}}^i_{context}$ is used for training the masked language modeling (MLM) task.

\subsection{Pre-training ERNIE-GeoL}
\subsubsection{Masked Language Modeling}
We use the whole word mask (WWM) strategy to make predictions for the phrases in each document. We use a query component analysis module deployed at Baidu Maps to split each document at the granularity of geographic entities.
Each geographic entity in a document has a 15\% probability of being masked and predicted by the language model during the training process.
For each word in the selected entity, we replace the word with a ``[MASK]'' token with 70\% probability, replace the word with a misspelled word with 10\% probability, replace the word with a random word with 10\% probability, and leave the word unchanged with 10\% probability.
The words in a query that do not match any words in the target POI name are treated as misspelled words.
With this training procedure, we can learn four types of toponym knowledge in the masked language modeling (MLM) task as follows.
(1) The natural language descriptions of POI name and address. 
(2) The relationships between POI name, address, and type. 
(3) The relationships between query, POI name, and address.
(4) The possible misspelling of POI name and address.

\subsubsection{Geocoding}
\label{sec:geocoding}
\begin{figure}[!t]
\setlength{\abovecaptionskip}{0.25cm}
	\centering
	\includegraphics[width=1.0\columnwidth,clip]{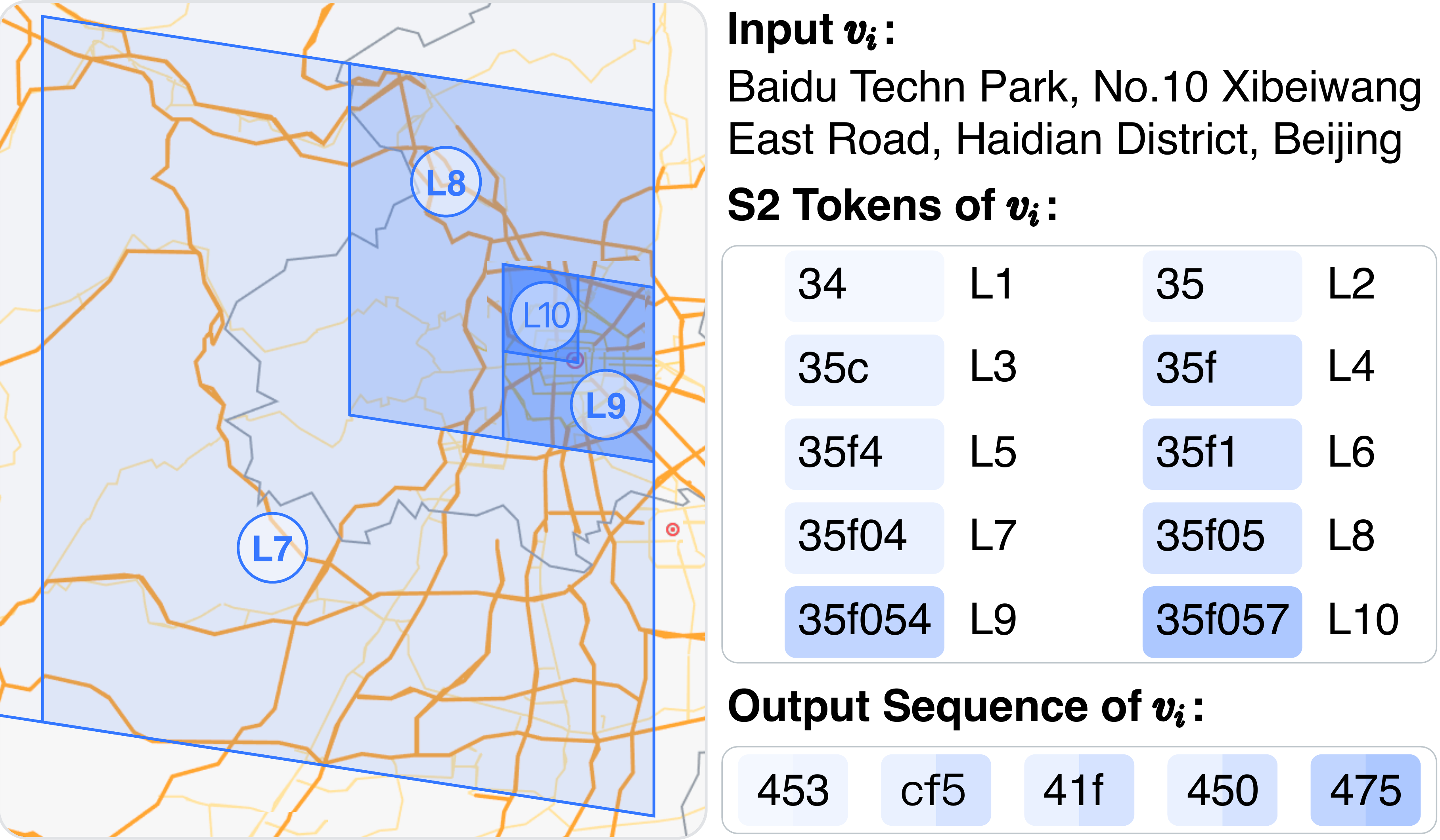}
	\caption{Illustration of the geocoding task. Each parallelogram represents an S2 cell of the corresponding level.}
	\label{fig:gc_task}
\vspace{-2mm}
\end{figure}

We provide new insights into learning the relations between text and geographic coordinates of a POI during pre-training.
Specifically, we adopt a feed-forward layer for each POI node to predict the IDs for multi-level S2 cells converted from the POI's coordinates.
In order to model the relations between text and coordinates in a more fine-grained manner, we set the highest level of S2 cell to 22, which covers an area approximate to 2 m$^2$.
However, in the granularity of level 22, the Earth's surface is divided into 105 trillion S2 cells.
Directly predicting these S2 cell's IDs will introduce an overwhelming number of output classes.
To address this problem, we first encode multi-level S2 tokens into a sequence, and then make predictions on it.
Figure \ref{fig:gc_task} shows an example of our proposed geocoding task, where the goal is to predict the sequence ``453 cf5 41f 450 475'' for the input text $v_i$.

For the same coordinates, the S2 tokens of two consecutive levels $2n+1$ and $2n+2$ ($n>0$) are designed to have a same length. 
Suppose their length is $T_i$, then the S2 tokens of the levels $2n-1$ and $2n$ would have a length of $T_i-1$.
Moreover, the tokens of the levels $2n - 1$ and $2n$ differ only in the last character.
To represent the multi-level S2 tokens of a POI as a single sequence efficiently and obtain the ability of parallel computing, we propose a new encoding/decoding scheme called 2Lt3C.
For simplicity, we use the term ``($2n$)'s token'' to denote ``the S2 token of the level $2n$''.
For encoding, we use the last character of ($2n-1$)'s token, the last character of ($2n$)'s token, and the shared penultimate character between ($2n-1$)'s and ($2n$)'s tokens, to generate a sub-sequence with three characters.
For example, the sub-sequence ``475'' is generated from the L9's token ``35f054'' and the L10's token ``35f057''.
Figure \ref{fig:gc_task} shows an example of the encoding scheme for levels from 1 to 10.
In this way, sufficient information of the multi-level S2 tokens can be encoded into a single sequence, which is also necessary to decode all levels.

Specifically, we cast the task of predicting the sequence encoded by 2Lt3C as $q$ independent classification problems, where each classifier learns to predict a character in this sequence.
Although this approach ignores potential correlations among characters, it offers the opportunity for parallel computing, and improves computational efficiency.
For each classifier, we use a softmax layer to compute the probability of classifying $v_i$ as a certain character by:
\begin{align}
    Pr(c_{i}|v_i) = \mathbf{softmax}(\widehat{\mathbf{h}}^i_{CLS} \mathbf{W}_v)  ,
\end{align}
where $\mathbf{W}_v \in \mathbb{R}^{d_n \times 16}$ is a trainable parameter, and $Pr(c_{i}|v_i) $ is the probability vector of a category $c_i \in \{0, \cdots, 15\}$ (6 English letters---a to f, and 10 Arabic numerals---0 to 9).
The number of classifiers ($q$) for S2 cells with level 22 is 33 determined by the design of 2Lt3C.

\begin{table*}[t]
\setlength{\abovecaptionskip}{0.12cm}
\setlength{\belowcaptionskip}{0.01cm}
\setlength{\tabcolsep}{5pt}
\caption{Tasks and Datasets used to evaluate ERNIE-GeoL. \#Dev = the number of samples in development set.}
    \begin{tabular}{@{}lcccccc@{}}
        \toprule
        \textbf{Task}  & \textbf{Problem Formulation}  & \textbf{Applicable Service} & \textbf{\#Train} & \textbf{\#Dev} & \textbf{\#Test} & \textbf{Metric} \\ \midrule
        Query intent classification & Sequence classification      & POI search engine       &   50,000    &  5,000   &   5,000   & Accuracy           \\
        Query-POI matching       & Sequence pair classification & POI search engine       & 140,614  &  4,000   &   4,000   & Accuracy           \\
        Address parsing          & Sequence labeling            & POI information processing  &  125,009     & 10,000    &   10,000   & Entity-level $F_1$    \\
        Geocoding               & Sequence classification      & POI information processing  &   2,171,114    &  1,000   &   1,000   & Acc@N km            \\
        Next POI recommendation  & Relevance ranking        & POI recommendation          &    6,398,231   &  300,000   &  300,000     & Acc@K               \\ \bottomrule
        \end{tabular}
    \label{tab:tasks}
\vspace{-2.5mm}
\end{table*}

\section{Experiments}
In this section, we present the results of ERNIE-GeoL on five geo-related tasks and the ablation experiments.

\subsection{Geo-related Tasks}
\subsubsection*{\textbf{Task \#1: Query Intent Classification}}
The query intent classification \cite{li2008learning} task aims at predicting the intent behind a query, which plays an important role in the POI search engine of Baidu Maps.
We define four intents, including the search for a specific POI, for a specific type of POI, for addresses, and for bus routes.

To evaluate its performance, we randomly sample 60,000 queries from the search logs of Baidu Maps and manually annotate them. 

In our experiments, we use $\mathbf{h}^i_{CLS}$ (see Equation~\ref{equ:1}) as the input representation, based on which a linear layer is adopted to perform the classification. We use accuracy as the evaluation metric, which represents the proportion of correctly predicted queries. 

\subsubsection*{\textbf{Task \#2: Query-POI Matching}}
The query-POI matching \cite{p3ac-kdd20,hgamn-kdd21} task aims at identifying the more relevant POI for a given query from a list of POIs.
There are four levels of relevance: the POI is exactly matched, highly relevant, weakly relevant, and irrelevant to the query.
The relevance score predicted by the matching model is an important feature for ranking the candidate POIs in the search engine of Baidu Maps.

To construct the dataset for this task, we randomly sample real-world queries from the search logs of Baidu Maps. 
For each query, we randomly sample 6 POIs from the top 10 POI candidates ranked by the POI search engine of Baidu Maps, as well as another four random POIs.
Then, we ask annotators to manually examine the 10 candidate POIs of each query and assign relevance levels to them.

In our experiments, we also use $\mathbf{h}^i_{CLS}$ to perform this task and employ accuracy as the evaluation metric. 

\subsubsection*{\textbf{Task \#3: Address Parsing}}
The task of address parsing~\cite{li2019neural} aims at parsing an address into a sequence of fine-grained geo-related chunks.
In this work, we designed 22 chunk types, among which 9 are for different levels of geographic areas, 2 for roads, 3 for different types of POI, 5 for details of POI location, and 3 for the auxiliary words that describe the location.
Plenty of modules in Baidu Maps, including a rule-based geocoding framework and the query understanding module of the POI search engine, rely on the output of the address parsing model.

To build the dataset for this task, we utilize two data sources.
One data source is the anonymized query logs of the geocoding service of Baidu Maps.
The other data source is the anonymized POI search logs of Baidu Maps.
Then, we ask the annotators to annotate every possible chunk of the addresses and queries.

We formulate this task as a sequence labeling problem and use ERNIE-GeoL + CRF as the model architecture.
We employ the entity-level $F_1$ score to evaluate the performance of chunk detection.

\subsubsection*{\textbf{Task \#4: Geocoding}}
\label{sec:geocoding_task}
Given a geo-located entity reference in text, the geocoding task \cite{goldberg2007text} aims at resolving the input to a corresponding location on the Earth.
As described in Section~\ref{sec:geocoding}, we directly predict S2 tokens for the input text.
The geocoding task is an essential service of mapping applications, its output is also a crucial feature required by other services like POI retrieval.

To collect the dataset for geocoding task, we use the same data built for the above-mentioned address parsing task.
For each address $ad_i$, we first obtain its geographic coordinates $gc_i$ predicted by an in-house geocoding service, and then correlate $ad_i$ with an S2 token converted from $gc_i$.
In this way, we can automatically construct large amounts of training data. 
Since the accuracy of our existing geocoding service cannot reach 100\%, we manually annotate 2,000 addresses as the development set and test set, respectively.
Different from the well-formatted address descriptions derived from our POI database for pre-training ERNIE-GeoL, the address descriptions in this dataset are generated by different users with varied knowledge and are not well formatted.
As such, the data leakage problem can be avoided.
Therefore, we can use this dataset to sufficiently validate a model's ability to correlate text with geographic coordinates.

The model architecture used for fine-tuning this task is the same as that described in Section~\ref{sec:geocoding}.
We use ``Accuracy@N km'' as the evaluation metric, which measures the percentage of predicted locations that are apart with a distance less than N km to their actual physical locations.
In our experiments, we set N to 3.

\subsubsection*{\textbf{Task \#5: Next POI Recommendation}}
Given a user's sequence of historical POI visits, the task of next POI recommendation (NPR) \cite{metapoirec-kdd21,yu2015survey} aims at recommending a list of POIs that the user is most likely to visit consequently.
NPR is an essential feature in Baidu Maps, which can help users explore new POIs.

To construct the dataset for NPR task, we use the anonymized POI visiting data in Beijing within a 6-month period from Baidu Maps.
For each POI sequence, we process a sliding window with a randomly selected width from 3 to 6 to get some sub-sequences, where the last POI is regarded as the label and the rest of POIs are taken as the historical visits.
For evaluation, we use 6 million POIs in Beijing as candidate POIs for retrieval.

In the fine-tuning phase, we use a two-tower approach similar to that used in our previous work~\cite{hgamn-kdd21}.
Taking ERNIE-GeoL as a feature encoder, we calculate the similarity between the graph-based visiting sequence representation and the target POI representation.
The key difference between pre-training and fine-tuning is that the input graph only contains POI-POI edges constructed with the POI visiting sequences.
In the evaluation phase, we first generate vectors for all the input sequences and candidate POIs. 
Then, we use the HNSW~\cite{malkov2018efficient} algorithm to process an approximate K-nearest neighbor search for retrieving the target POI.
We use Acc@K as the evaluation metric, which calculates the proportion of the recommended POI sequences where the visited POI appears within the top K positions.
In our experiments, we set K to 50.

The five tasks and datasets used to evaluate ERNIE-GeoL are summarized in Table \ref{tab:tasks}.

\subsection{Experimental Setup}
\subsubsection{Datasets}
In our experiments, we construct the heterogeneous graph using search logs within a 3-month period from Baidu Maps.
The heterogeneous graph contains 40 million POI nodes, 120 million query nodes, 175 million Query-click-POI edges, 1,574 million Origin-to-Destination edges, and 363 million POI-(co-locate with)-POI edges.
We use the random walk algorithm on the heterogeneous graph to sample a sequence of nodes as an input document.
The sampling weights of Query-click-POI, Origin-to-Destination, and POI-(co-locate with)-POI edges are set as $\lambda_1=0.5$, $\lambda_2=0.25$, and $\lambda_3=0.25$, respectively.
We sample 800 million documents from the graph, which contain 400 billion words. 
Each document contains an average of 10 nodes.

\begin{table*}[htbp]
\setlength{\abovecaptionskip}{0.12cm}
\setlength{\belowcaptionskip}{0.01cm}
\caption{Comparison of pre-trained models on five geo-related tasks. Average means the averaged score of five tasks.}
    \begin{tabular}{lcccccc}
    \toprule
      \textbf{Pre-trained Model} &
      \makecell{\textbf{Query Intent} \\ \textbf{Classification}} &
      \makecell{\textbf{Query-POI} \\ \textbf{Matching}} &
      \textbf{Address Parsing} &
      \textbf{Geocoding} &
      \makecell{\textbf{Next POI} \\ \textbf{Recommendation}} &
      \textbf{Average} \\ \midrule \midrule
      \textbf{BERT}~\cite{Devlin2019BERTPO} &
      0.8875 &
      0.8279 &
      0.8452 &
      0.4592 &
      0.1092 &
      0.6258 \\
      \textbf{RoBERTa}~\cite{Liu2019RoBERTaAR} &
      0.8907 &
      0.8285 &
      0.8497 &
      0.4618 &
      0.1115 &
      0.6284 \\
      \textbf{ERNIE 2.0}~\cite{sun2020ernie} &
      0.8919 &
      0.8290 &
      0.8511 &
      0.4636 &
      0.1198 &
      0.6311 \\ \midrule
      \textbf{ERNIE-GeoL} &
      \textbf{0.9161} &
      \textbf{0.8332} &
      \textbf{0.8794} &
      \textbf{0.6545} &
      \textbf{0.1556} &
      \textbf{0.6878} \\
      \quad - w/o geocoding task &
      0.9068 &
      0.8050 &
      0.8682 &
      0.5436 &
      0.1301 &
      0.6507 \\
      \quad - w/o heterogeneous graph &
      0.9101 &
      0.8025 &
      0.8688 &
      0.5809 &
      0.1359 &
      0.6596 \\
      \quad - w/o O-t-D edge &
      0.9076 &
      0.8129 &
      0.8715 &
      0.6273 &
      0.1403 &
      0.6719 \\
      \quad - w/o Q-c-P edge &
      0.9155 &
      0.8072 &
      0.8784 &
      0.6381 &
      0.1413 &
      0.6761 \\
      \quad - w/o P-c-P edge &
      0.9148 &
      0.8305 &
      0.8780 &
      0.6164 &
      0.1458 &
      0.6771 \\\bottomrule
    \end{tabular}
    \label{tab:main_results}
\vspace{-2.5mm}
\end{table*}

\subsubsection{Baselines}
We evaluate ERNIE-GeoL against three strong generic PTMs as follows:
\begin{itemize}
    \item \textbf{BERT}~\cite{Devlin2019BERTPO} is a Transformer-based pre-trained language model, which has made impressive gains on many NLP tasks.
    \item \textbf{RoBERTa}~\cite{Liu2019RoBERTaAR} is a variant of BERT with enhanced strategies, which improves the performance on several NLP tasks.
    \item \textbf{ERNIE 2.0}~\cite{sun2020ernie} is another variant of BERT, which facilitates continuous learning of multiple pre-training tasks. It outperforms BERT and RoBERTa on many Chinese NLP tasks.
\end{itemize}

We also perform ablation experiments over a number of facets of ERNIE-GeoL to figure out their relative importance, which include:
\begin{itemize}
    \item \textbf{ERNIE-GeoL} is the complete model depicted in Section \ref{sec:methods}.
    \item \textbf{ERNIE-GeoL w/o geocoding task.} In this setting, we remove the pre-training task of geocoding (see Section \ref{sec:geocoding}).
    \item \textbf{ERNIE-GeoL w/o heterogeneous graph.} In this setting, we remove all edges in the heterogeneous graph and pre-train ERNIE-GeoL using the text representation of each node, as described in Section \ref{sec:node_text_repr}.
    \item \textbf{ERNIE-GeoL w/o a specific type of edge.} In this group of settings, we remove the edge of Origin-to-Destination (O-t-D), query-click-POI (Q-c-P), and POI-(co-locate with)-POI (P-c-P) when constructing the heterogeneous graph, respectively.
\end{itemize}

All the pre-training and fine-tuning procedures are implemented using the PaddlePaddle deep learning framework.
We use Adam optimizer \cite{kingma2014adam}, with the learning rate initialized to $5 \times 10^{-5}$ and gradually decreased during the process of training.
The hyper-parameters of all PTMs are the same as those used in BERT\_{BASE} \cite{Devlin2019BERTPO} (number of hidden layers = 12, hidden layer size = 768, number of attention heads = 12, number of total parameters = 110M).
The training takes about one week on 16 Nvidia A100 GPUs.

\subsection{Results and Analysis}
Table~\ref{tab:main_results} shows the main experiment results.
The last column ``Average'' presents the averaged score of the five tasks.

\subsubsection{Overall Performance}
We first evaluate whether ERNIE-GeoL can improve the performance of the five geo-related tasks. 
Before ERNIE-GeoL, we have applied ERNIE to improve these tasks at Baidu Maps and obtained initial gains.
However, a clear gain plateau over time was observed due to the lack of geographic domain knowledge in ERNIE.
This motivated us to design and develop a geography-and-language pre-trained model ERNIE-GeoL for improving the geo-related tasks at Baidu Maps.
The results in Table \ref{tab:main_results} show that ERNIE-GeoL significantly outperforms all three generic PTMs (i.e., BERT, RoBERTa, and ERNIE) by a large margin, and achieves a highest average score of 0.6878.
This demonstrates that our model is more effective in dealing with geo-realted tasks.
One of the main reasons for this superiority is that ERNIE-GeoL has comprehensively learned the geographic knowledge.

\subsubsection{Ablation Studies}
We perform ablation experiments to understand the relative importance of different facets of ERNIE-GeoL.

First, we study the effect of pre-training tasks.
Compared with ERNIE-GeoL, the average score of ``ERNIE-GeoL w/o geocoding task'' dramatically decreases by an absolute 3.71\%, which is the largest drop observed among all ablation experiments.
This demonstrates the impact brought by the geocoding task.
The main reason is that geo-information plays an vital role in geo-related tasks.
Therefore, the ability to learn a universal representation of geography-language is crucial for pre-training a geographic model.

Second, we evaluate the effect of heterogeneous graph.
The results in Table~\ref{tab:main_results} show that removing the graph (ERNIE-GeoL w/o heterogeneous graph) hurts performance significantly on all tasks.
This demonstrates the significance of graph structure in the training data, which can effectively integrate spatial knowledge with text.

Third, we examine the impact of different edge types.
The results show that removing individual edges (ERNIE-GeoL w/o P-c-P edge, w/o O-t-D edge, and w/o Q-c-P edge) hurts performance on all tasks.
This indicates that all types of edges are essential for pre-training a geographic model, and they can work together as complements.
Removing the O-t-D edge (ERNIE-GeoL w/o O-t-D edge) leads to the largest drops among the three ablations of edges, this demonstrates the importance of human mobility data in geo-related tasks.

\begin{figure*}[!th]
\setlength{\abovecaptionskip}{0.15cm}
  \begin{subfigure}[b]{0.45\textwidth}
    \setlength{\abovecaptionskip}{0.05cm} 
    \setlength{\belowcaptionskip}{0.05cm}    \includegraphics[width=\textwidth]{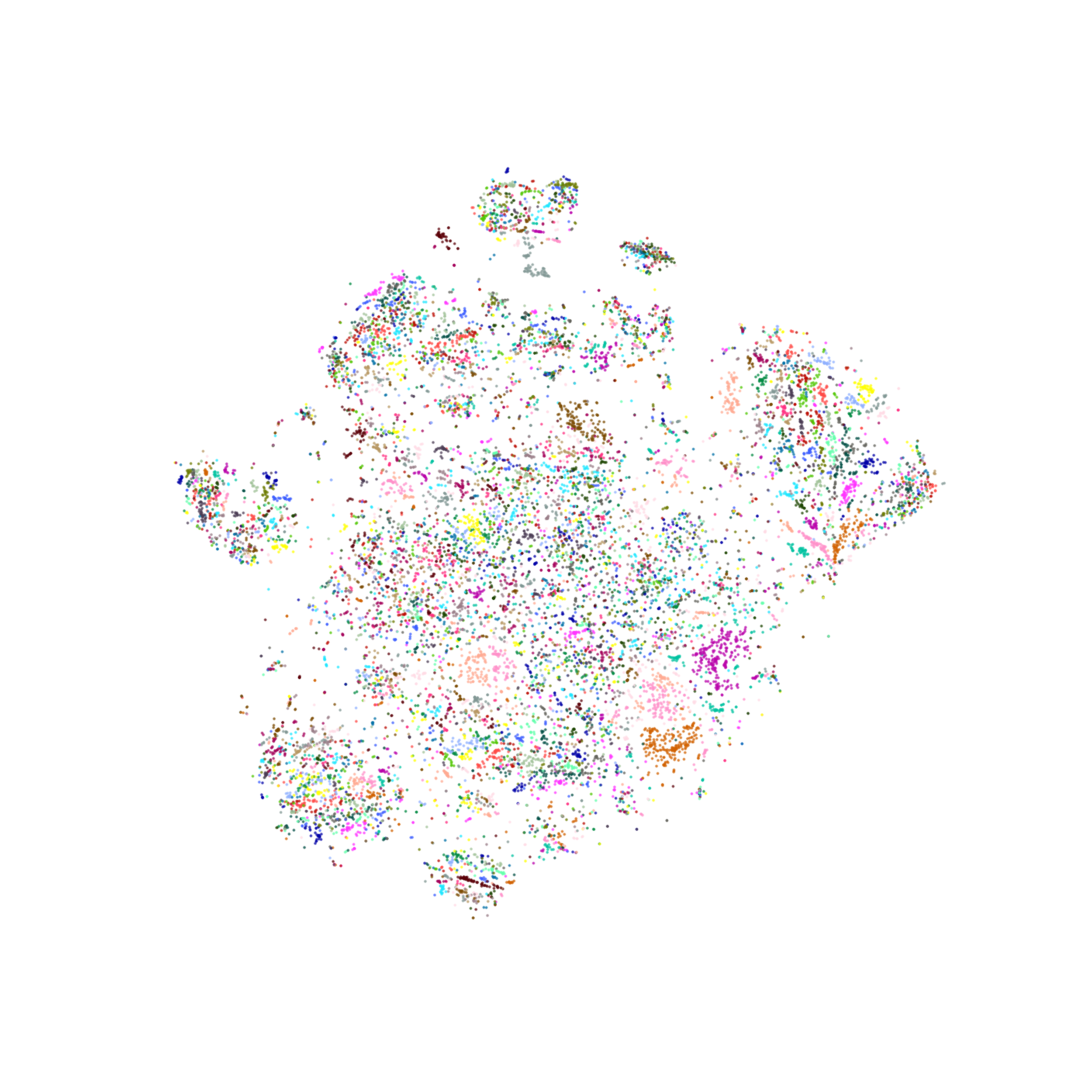}
    \caption{The t-SNE visualization of embeddings produced by BERT.}
    \label{fig:proj_bert}
  \end{subfigure}
  \begin{subfigure}[b]{0.53\textwidth}
    \setlength{\abovecaptionskip}{0.05cm} 
    \setlength{\belowcaptionskip}{0.05cm}   
    \includegraphics[width=\textwidth]{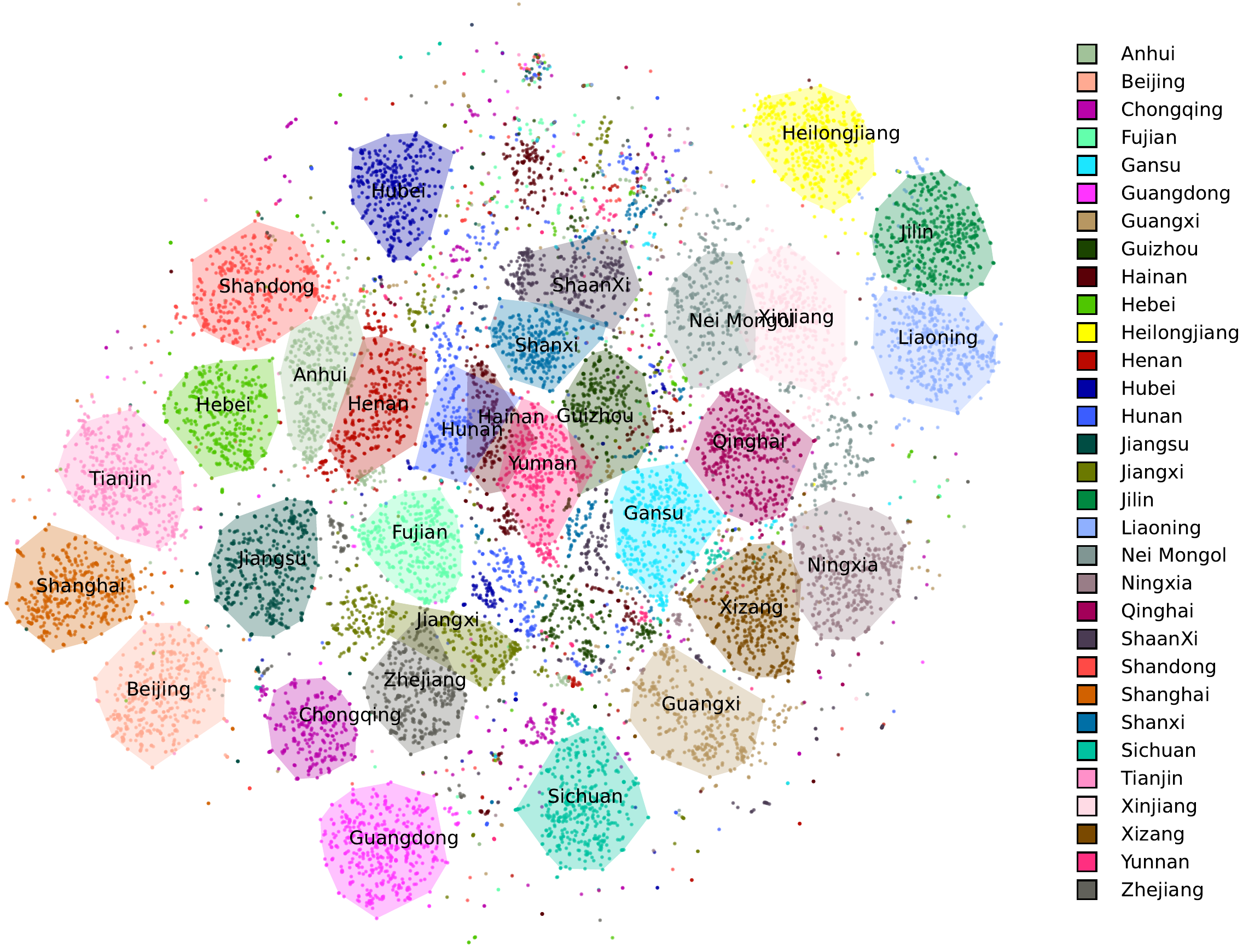}
    \caption{The t-SNE visualization of embeddings produced by ERNIE-GeoL.}
    \label{fig:proj_polaris}
  \end{subfigure}
\vspace{-1mm}
\caption{A 2D t-SNE projection of top 500 searched POIs in 31 provinces excluding Hong Kong, Macao, and Taiwan of China.}
\label{fig:poi_projection}
\vspace{-3mm}
\end{figure*}

Finally, we evaluate the performance of individual tasks.
(1) Compared with ERNIE 2.0 (the best performing generic PTM), the performance of geocoding trained with ERNIE-GeoL achieves the largest increase of 19.09\% (by absolute value).
We also showcase how this model can robustly deal with different types of input text in Appendix \ref{appendix:geocoding_case}.
Moreover, the ``ERNIE-GeoL w/o geocoding task'' model performs worse than the ERNIE-GeoL model on the geocoding task, with a large drop of 11.09\% (by absolute value).
The main reason is that the performance of a geocoding model heavily relies on its ability to correlate text with geographic coordinates.
This demonstrates that ERNIE-GeoL has learned sufficient geographic knowledge about text and coordinates during pre-training.
(2) The task that has the second highest benefits among five tasks is next POI recommendation, with an absolute improvement of 3.58\% over ERNIE.
Moreover, removing the O-t-D edge (ERNIE-GeoL w/o O-t-D edge) leads to a drop of 1.53\% on the next POI recommendation task.
The main reason is that prior knowledge of the distribution of human mobility data is crucial for the next POI recommendation task.
This indicates that ERNIE-GeoL has learned the distribution of human mobility from the training data.
(3) Among the generic PTMs, ERNIE 2.0 outperforms BERT and RoBERTa on five tasks. This shows that the optimization made by ERNIE 2.0 for dealing with Chinese NLP tasks, such as the masking strategy that masks phrases and entities rather than individual sub-words, can be a benefit for Chinese toponym masking.

\subsection{A Qualitative Study on Geo-Linguistic Knowledge ERNIE-GeoL Has Learned}
\subsubsection{Embedding Projection.}
For an intuitive understanding of the geo-linguistic knowledge ERNIE-GeoL has learned, we encode POIs to embeddings and project them into a two-dimensional space using the t-distributed stochastic neighbor embedding (t-SNE) \cite{tsne} method.
We select the top 500 frequently searched POIs at Baidu Maps in 31 provinces excluding Hong Kong, Macao, and Taiwan of China.
To get each POI's embedding, we first concatenate its name and address as the input for PTMs.
Then, we use the output hidden state of the ``[CLS]'' token as its embedding.
Figure \ref{fig:poi_projection} illustrates the t-SNE visualization of embeddings predicted by BERT and ERNIE-GeoL.
From which, we can clearly see that the t-SNE embeddings predicted by ERNIE-GeoL (Figure \ref{fig:proj_polaris}) are more discriminative than those predicted by BERT (Figure \ref{fig:proj_bert}), which shows that POIs located in different provinces can be geographically differentiated by our ERNIE-GeoL model.
Moreover, in Figure~\ref{fig:proj_polaris}, geographically adjacent provinces tend to be adjacent to one another.
For example, the points of Heilongjiang, Jilin, and Liaoning provinces co-locate in the upper-right corner of Figure~\ref{fig:proj_polaris}.
In reality, the three provinces are also adjacent to each other in the northeast part of China.
These observations show that ERNIE-GeoL has successfully learned the relations between text and their real-world geographic locations.

\begin{figure}[htbp]
  \setlength{\abovecaptionskip}{0.15cm}
    \begin{subfigure}[b]{0.235\textwidth}
      \includegraphics[width=\textwidth]{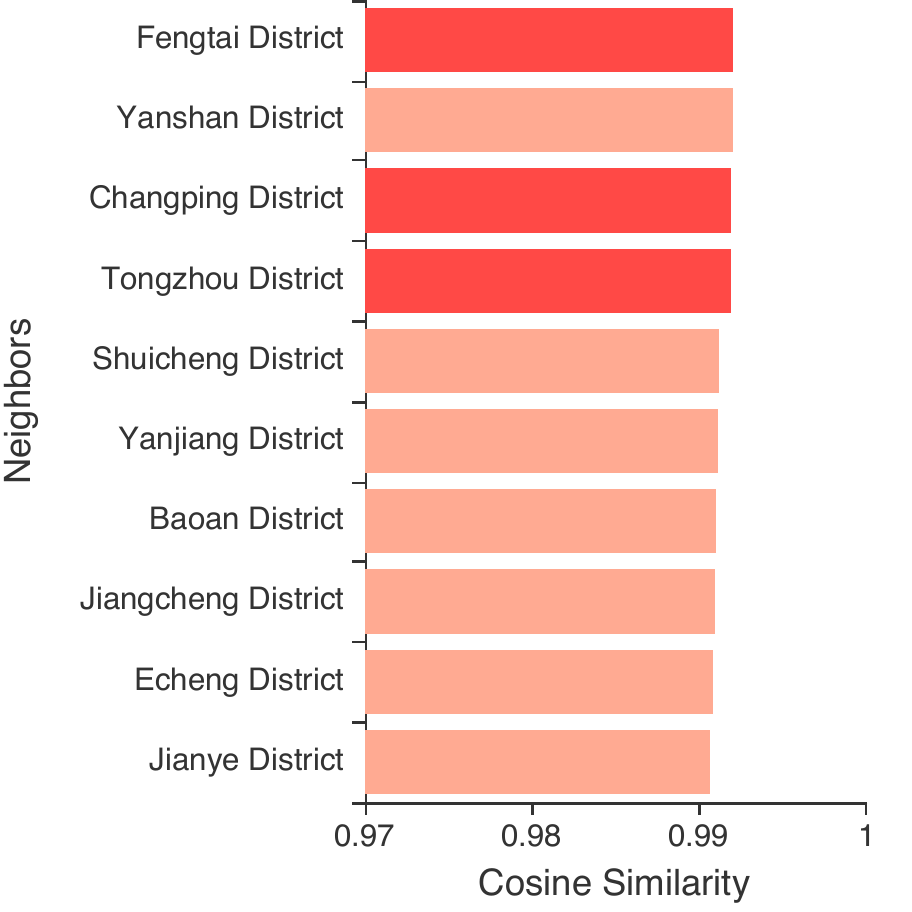}
      \vspace{-5.8mm}
      \caption{Results of BERT.}
      \label{fig:relation_area_bert}
    \end{subfigure}
    \begin{subfigure}[b]{0.235\textwidth}
      \includegraphics[width=\textwidth]{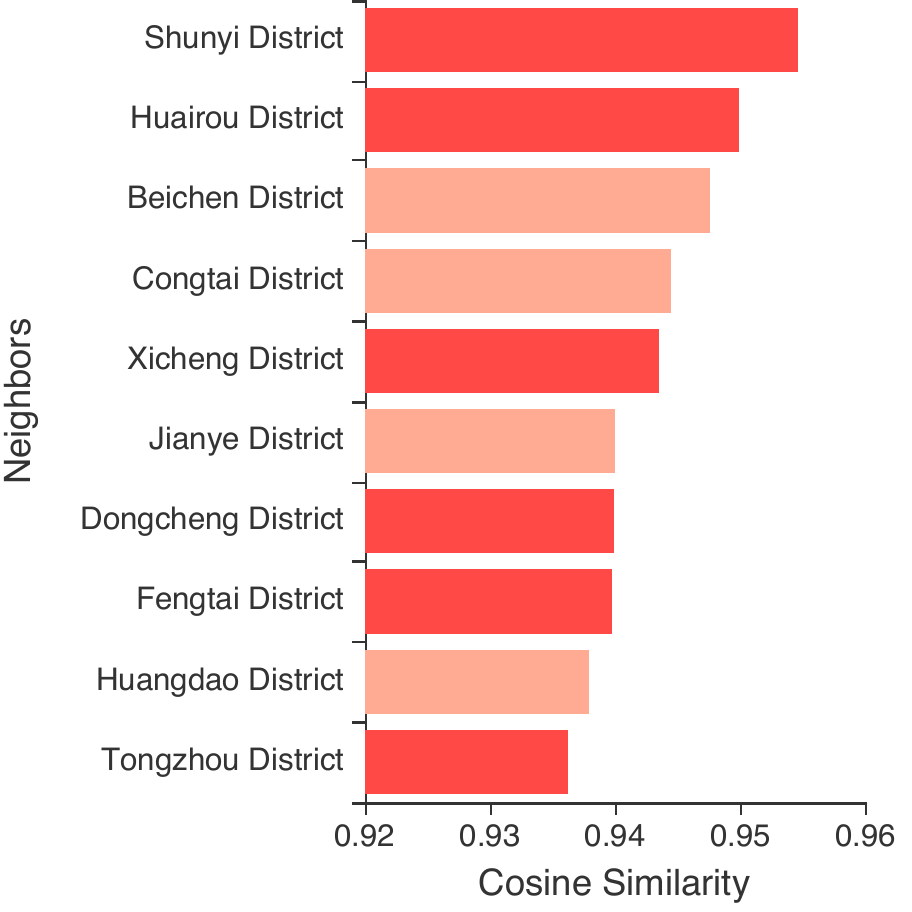}
      \vspace{-5.8mm}
      \caption{Results of ERNIE-GeoL.}
      \label{fig:relation_area_polaris}
    \end{subfigure}

\vspace{-1mm}
\caption{Nearest Neighbors of ``Huangpu District - Shanghai + Beijing''. The darker bar represents the district of Beijing.}
\label{fig:relation_area}
\vspace{-2mm}
\end{figure}
  
\begin{figure}[htbp]
\setlength{\abovecaptionskip}{0.15cm}
    \begin{subfigure}[b]{0.235\textwidth}
      \includegraphics[width=\textwidth]{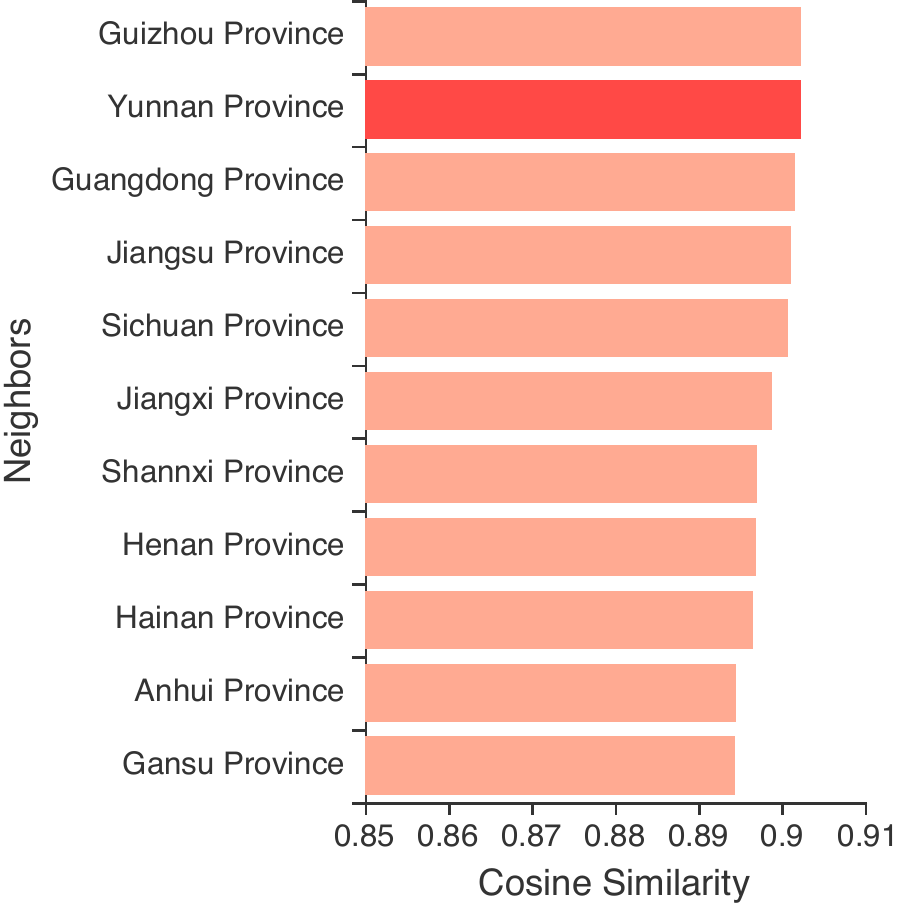}
      \vspace{-5.8mm}
      \caption{Results of BERT.}
      \label{fig:relation_prov_bert}
    \end{subfigure}
    \begin{subfigure}[b]{0.235\textwidth}
      \includegraphics[width=\textwidth]{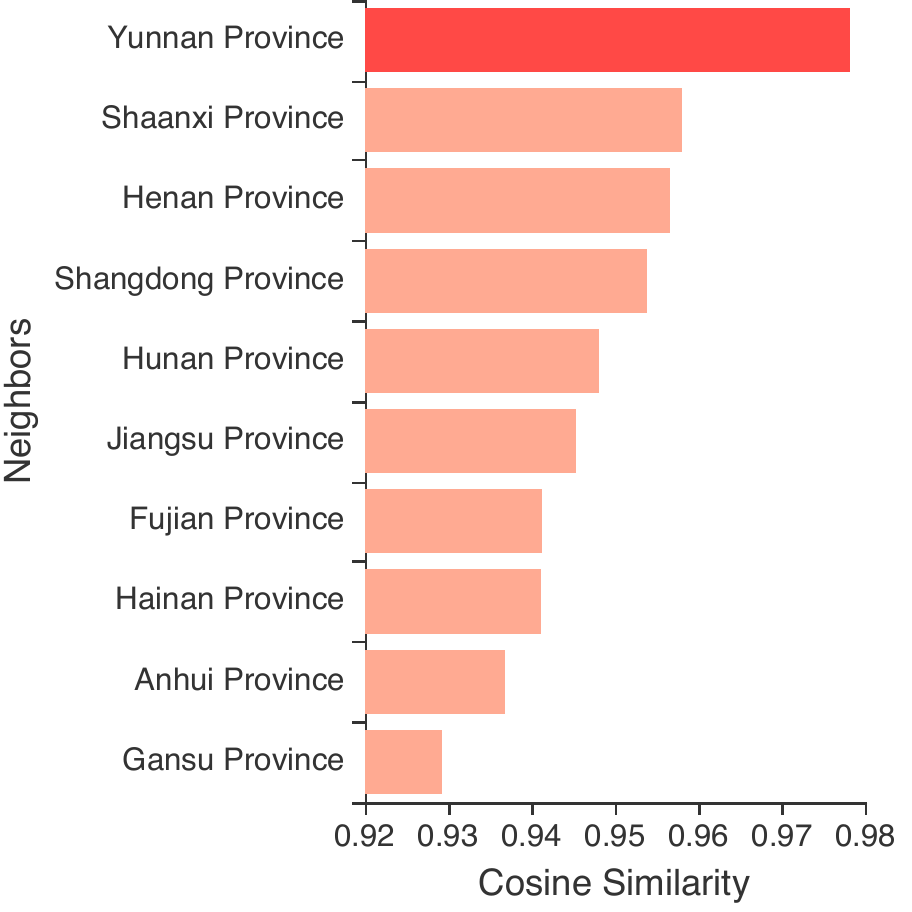}
      \vspace{-5.8mm}
      \caption{Results of ERNIE-GeoL.}
      \label{fig:relation_prov_polaris}
    \end{subfigure}

\vspace{-1mm}
\caption{Nearest Neighbors of ``Guangdong Province - Guangzhou + Kunming''. The darker bar represents Yunnan Province, the capital city of which is Kunming.}
\label{fig:relation_prov}
\vspace{-4mm}
\end{figure}

\subsubsection{Geographic Analogy}
Following the famous $king-man+woman \thickapprox queen$ example \cite{mikolov2013linguistic}, we curate two test cases to examine how well ERNIE-GeoL can learn the geographic analogy.
As shown in Figure~\ref{fig:relation_area} and Figure~\ref{fig:relation_prov}, we test the geographic analogy of ``district of a city'' and ``capital city of a province'', respectively.
We show the top 10 neighbors ranked by their cosine similarity to the query.
In Figure~\ref{fig:relation_area}, the query is set to ``Huangpu District - Shanghai + Beijing'', and the candidate neighbors are set to district names of all Chinese cities.
We can observe that ERNIE-GeoL recalls more Beijing's districts than BERT.
In Figure~\ref{fig:relation_prov}, the query is set to ``Guangdong Province - Guangzhou + Kunming'', and the candidate neighbors are set to all Chinese province names.
We can observe that ERNIE-GeoL recalls the target neighbor, ``Yunnan Province'', with the highest score.
Moreover, in both Figures, the cosine similarity predicted by BERT is less discriminative than that by ERNIE-GeoL.
Such observations show that ERNIE-GeoL has learned the spatial relationships between different geo-located entities.

\section{Related Work}
Here we briefly review the closely related work in the fields of domain-specific PTMs and PTMs utilizing multi-source data.

\subsection{Domain-specific PTMs}
Existing domain-specific PTMs mainly lie in the domain of healthcare~\cite{Huang2019ClinicalBERTMC,alsentzer2019publicly}, biomedical~\cite{lee2020biobert,gu2021domain,fang2022geometry}, and academic \& research~\cite{beltagy2019scibert}.
Most of them learn the domain-specific knowledge by pre-training on domain-specific corpora with the MLM pre-training task.
The most relevant work to ours is OAG-BERT~\cite{Liu2021OAGBERTPH}, which is an academic PTM pre-trained using the heterogeneous knowledge from an academic knowledge graph.
However, they do not model the graph structure explicitly, like our proposed method.

\subsection{PTMs with Multi-source Data}
Most existing multimodal PTMs are designed to model the relations between text and image~\cite{lu2019vilbert}, video~\cite{sun2019videobert}, and audio~\cite{Chuang2019SpeechBERTCP}.
However, pre-training a geographic PTM requires modeling the relations between text and geographic coordinates (in the form of numerical data).
Such an intersection of multiple modalities of text and numbers has not been well explored in the literature.

\section{Conclusions and Future Work}
This paper presents an industrial solution for building a geography-and-language pre-trained model that has already been deployed at Baidu Maps.
We propose a framework, named ERNIE-GeoL, that comprehensively learns geographic domain knowledge.
Sampled from a heterogeneous graph constructed upon the POI database and the search logs of Baidu Maps, the documents used for pre-training ERNIE-GeoL are injected with toponym and spatial knowledge.
The backbone network of ERNIE-GeoL contains an aggregation layer for modeling the graph structure entailed in the input documents.
ERNIE-GeoL adopts two pre-training objectives, including masked language modeling and geocoding, for guiding the model to learn the toponym and spatial knowledge, respectively. 
We evaluate ERNIE-GeoL on a benchmark built based on five tasks that provide fundamental support for an essential mapping service.
The experiment results and ablation studies show that ERNIE-GeoL outperforms previous generic pre-trained models, demonstrating that ERNIE-GeoL can serve as a promising foundation for a wide range of geo-related tasks.

% In this paper, the geographic data used for training ERNIE-GeoL mainly focus on the geographic features of POIs and how users interact with the POIs.
In future work, to make ERNIE-GeoL capable of handling a more wide range of geographic applications, we plan to enhance ERNIE-GeoL with satellite images and street views.

% \begin{acks}
% 
% \end{acks}

\balance 
\bibliographystyle{ACM-Reference-Format}
\bibliography{main}

\clearpage
% \nobalance

\appendix

\section{Appendix} 
In this section, we first compare different DGG systems, and detail the node representation method.
Then, we showcase an example of the generated pre-training data, and how ERNIE-GeoL can help the geocoding model to effectively handle different types of input.
We also show how to fine-tune downstream tasks with ERNIE-GeoL.

\subsection{Comparison of Different DGG Systems}
\label{appendix:geocode_system}
Table \ref{table:compare_ddg} shows the comparison of three recently proposed DGG systems including Geohash\footnote{\url{https://en.wikipedia.org/wiki/Geohash}}, H3 system\footnote{\url{https://eng.uber.com/h3/}}, and S2 geometry\footnote{\url{https://s2geometry.io}}.
We compare them in terms of projection method, geographic containment, and the number of levels.

\textbf{First}, modern DGG systems usually adopt a projection method to transform a three-dimensional location on the Earth into a two-dimensional point on a map. Compared with Geohash, H3 and S2 use spherical projections, which significantly reduce the distortion brought by the Mercator projection.
\textbf{Second}, all three systems are hierarchical systems in which each cell is a ``box reference'' to a subset of cells.
Here, we use ``geographic containment'' to indicate whether all child cells are perfectly contained within a parent cell. Compared with Geohash and S2, H3's geographic containment is approximate since it uses hexagons for tilling.
\textbf{Third}, the more levels a system supports, the greater is its flexibility. S2 supports 31 levels of hierarchy, while Geohash/H3 only supports 12/15.

\begin{table}[h]
\vspace{-0.5mm}
\setlength{\abovecaptionskip}{0.15cm}
  \caption{Comparison of three DGG systems.}
  \begin{tabular}{@{}lccc@{}}
    \toprule
    \makecell{\textbf{Geocode} \\ \textbf{System}}  & \makecell{\textbf{Projection} \\ \textbf{Method}}        & \makecell{\textbf{Geographic} \\ \textbf{Containment}}  & \textbf{\#Levels}  \\ \midrule
    Geohash        & Mercator     & Accurate                 & 12            \\
    H3             & Icosahedron & Approximate                & 15        \\ 
    S2            & Hexahedron  & Accurate                 & 31      \\ \bottomrule
    \label{table:compare_ddg}
    \end{tabular}
\vspace{-3.5mm}
\end{table}

We choose S2 as our DGG system based on the following three considerations.
First, it uses the hexahedral projection scheme to avoid distortion, which is able to preserve the correct topology of the Earth.
Second, S2 supports 31 levels of hierarchy, which offers the best flexibility.
Third, S2's geographic containment is accurate, which underpins the design, development, and implementation of the proposed geocoding pre-training task.

\subsection{Node Representation}
\label{appendix:qp_repr}
As described in Section \ref{sec:node_text_repr}, we uniformly use the textual information of each node to represent the query nodes and POI nodes.

For a query node, since the query used to find a user's desired POI is already in text format, we use its text to represent the query node. 
For example, in Figure \ref{fig:random_wak_sample}, the text ``\textit{No.1 Songzhuang Road Yichi Food}'' is used to represent the query node ``Q1''.

For a POI node, we first extract the POI's text fields of \textbf{name}, \textbf{address}, and \textbf{type} from the POI database of Baidu Maps.
Then, we generate a text sequence by joining the three fields with a special token ``[SEP]'', to represent the POI node.
For example, in Figure \ref{fig:random_wak_sample}, the text sequence ``\textit{Yizi Food (Suzhou) Co. [SEP] No.1, Songxiang, Suzhou Industrial Park, Suzhou, Jiangsu Province [SEP] Company}'' is used to represent the POI node ``01'', where ``\textit{Yizi Food (Suzhou) Co.}'', ``\textit{No.1, Songxiang, Suzhou Industrial Park, Suzhou, Jiangsu Province}'', and ``\textit{Company}'' are the name, address, and type of POI ``01'', respectively.

\subsection{A Case Study on the Pre-training Data}
\label{appendix:data}
\begin{figure*}[!ht]
\setlength{\abovecaptionskip}{0.2cm}
\centering
  \includegraphics[width=\textwidth,clip]{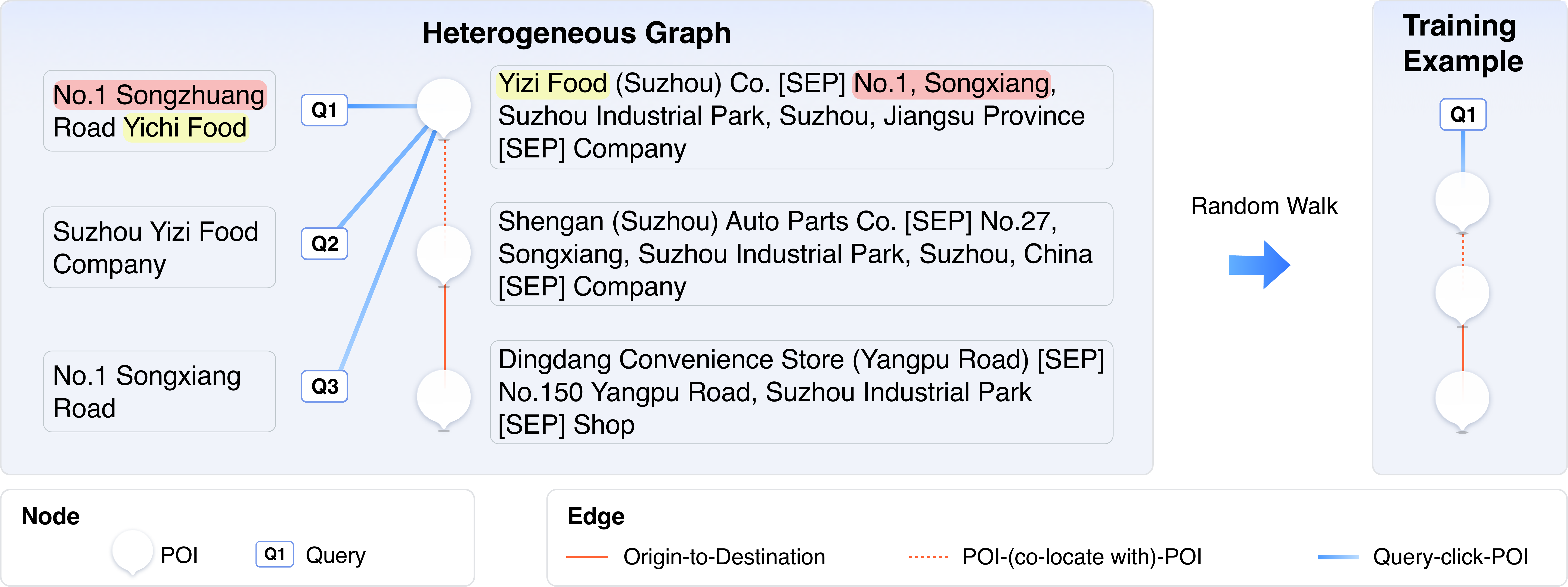}
  \caption{An example of the heterogeneous graph and a training example generated from the graph.}
  \label{fig:random_wak_sample}
\end{figure*}

Figure \ref{fig:random_wak_sample} showcases an example of the heterogeneous graph and a training example generated from the graph.
We show how toponym knowledge and spatial knowledge (see Section \ref{sec:intro} for details) are entailed by the generated training example.

For toponym knowledge, we can see that a POI's text sequence contains the exact toponyms like the POI name and the address that the POI locates in.
In addition, the queries often contain different formulations of the same toponym.
On the one hand, different formulations can reveal multiple aliases of the same POI, which can help the model learn diversified expressions of a formal toponym.
For example, we can observe from Figure \ref{fig:random_wak_sample} that users have used three different queries ``Q1'', ``Q2'', and ``Q3'' to search the same POI ``01''.
On the other hand, large-scale formulations uncover the majority of the frequent misspelling patterns, from which we can learn misspelling regularities that make the pre-training model to robustly generalize to unseen misspelling patterns.
For example, the yellow-colored formulation ``\textit{Yichi Food}'' is a misspelled query of the POI name ``\textit{Yizi Food}'' as shown in Figure \ref{fig:random_wak_sample}.

For spatial knowledge, we can observe from Figure \ref{fig:random_wak_sample} that the graph contains human mobility patterns.
For example, the POI visit sequence ``02-to-03'' represents a mobility pattern of ``workplace-to-market''.
These patterns are valuable for geo-related tasks that need to take human mobility into consideration, such as the next POI recommendation task.
Moreover, we can see that the graph also contains the spatial correlation information.
For example, the POIs ``01'' and ``02'' are linked by a POI-(co-located with)-POI edge.

\subsection{Typical Examples of the Geocoding Task}
\label{appendix:geocoding_case}
\begin{figure*}[!ht]
\setlength{\abovecaptionskip}{0.15cm}
\centering
    \includegraphics[width=\textwidth,trim={0.1cm 0.2cm 0.1cm 0.0cm},clip]{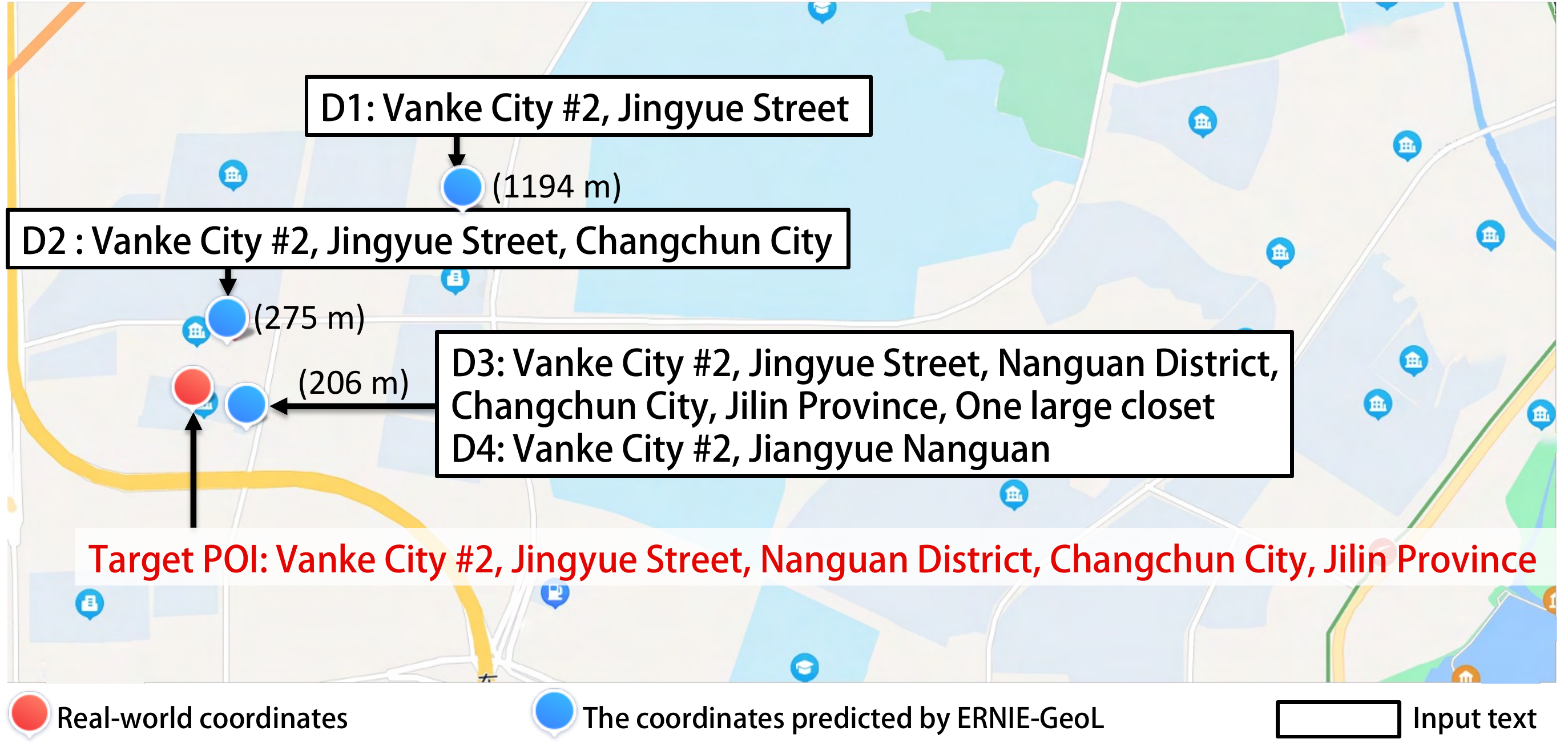}
    \caption{Four typical examples of the geocoding task.}
    \label{fig:geo_coding_sample}
\vspace{-2mm}
\end{figure*}

As described in Section \ref{sec:geocoding_task},
given a geo-located entity reference in text as input, the geocoding task aims at resolving the input to the corresponding location on the Earth.
Figure \ref{fig:geo_coding_sample} illustrates four typical examples of the geocoding task.
Given four different input text (D1, D2, D3, and D4) that attempt to describe the target POI (\textit{Vanke City \#2, Jingyue Street, Nanguan District, Changchun City, Jilin Province}), the coordinates predicted by our model are represented as blue pins in the figure.
First, we use the ERNIE-GeoL enhanced geocoding model to generate an S2 token for each input text.
Then, we take as output the coordinates located in the centroid of the territory represented by the generated S2 token.

We can observe from Figure \ref{fig:geo_coding_sample} that our model can make accurate predictions for different variety of descriptions of the same POI.
These examples show that our model is able to robustly handle the following three categories of descriptions. 
(1) \textbf{Incomplete description}. In this category, the description of an address usually omits some essential address elements. For example, in the input D1 (\textit{Vanke City \#2, Jingyue Street}), the descriptions of province and city are both omitted.
(2) \textbf{Informal description.} In this category, some address elements are described informally. For example, in the input D4, the street and district name ``\textit{Jingyue Street, Nanguan District}'' is informally abbreviated as ``\textit{Jingyue Nanguan}''.
(3) \textbf{Geographically irrelevant description.} In this category, the description usually contains some elements that are irrelevant to geographic information. For example, the ``\textit{One large closet}'' in input D3 is totally irrelevant to the POI. In our practice, such descriptions often confuse the model and lead to inaccurate predictions.

\subsection{How to Fine-tune Tasks with ERNIE-GeoL}
In general, we can use ERNIE-GeoL as a feature encoder for fine-tuning downstream tasks. There are two options for encoding the input of the downstream tasks.

(1) If the input of the downstream tasks contains a graph structure, we can encode each node $v_i$ in the input as $\{\widehat{\mathbf{h}}^i_{CLS}$, $\widehat{\mathbf{H}}^i_{context}\}$ calculated by the Equation \ref{equ:2}. In our practice, typical tasks that have an input with a graph structure include POI recommendation, context-aware POI retrieval, and task-oriented dialogue generation in our intelligent voice assistant.

(2) If the input of the downstream tasks is a text sequence, we can regard the input as a single node and encode it as $\{\mathbf{h}^i_{CLS}, \mathbf{H}^i_{context}\}$ calculated by the Equation \ref{equ:1}.
For example, the tasks that only have a text sequence as input include context-insensitive query intent classification, address parsing, and geocoding.

Then, the encoded input can be further handled by the task-specific neural network layers to obtain the final output for training or inferencing.

\end{document}